\documentclass[11pt]{article}
\pdfoutput=1

\usepackage[]{EMNLP2022}

\usepackage{times}
\usepackage{latexsym}
\usepackage{subcaption}
\usepackage{graphicx}
\usepackage{amsmath}
\usepackage{amssymb}
\usepackage{graphicx}
\usepackage{multirow}
\usepackage{makecell}
\usepackage{microtype}
\usepackage{algorithm}
\usepackage{algorithmicx}
\usepackage{algpseudocode}
\usepackage{hyperref}

\usepackage[extralight]{FiraSans}
\usepackage[T1]{fontenc}

\usepackage[utf8]{inputenc}

\usepackage{microtype}

\usepackage{inconsolata}

\title{Efficient Pre-training of Masked Language Model \\
via Concept-based Curriculum Masking}

\author{Mingyu Lee$^{1}$\thanks{\ \ These authors contributed equally to this work.}    , \: 
  Jun-Hyung Park$^{2*}$, \: 
  \textbf{Junho Kim}$^1$, \:
  \textbf{Kang-Min Kim}$^3$, \: 
  \textbf{SangKeun Lee}$^{1, 2}$ \\
  $^1$Department of Artificial Intelligence\ \  $^2$Department of Computer Science and Engineering \\
  Korea University, Seoul, Republic of Korea\\
  $^3$Department of Data Science, The Catholic University of Korea, Bucheon, Republic of Korea \\ 
  \texttt{\{decon9201, irish07, monocrat\}@korea.ac.kr} \\
  \texttt{kangmin89@catholic.ac.kr}, \ \  \texttt{yalphy@korea.ac.kr}\\ 
  }

\begin{document}
\maketitle
\begin{abstract}
Masked language modeling (MLM) has been widely used for pre-training effective bidirectional representations, but incurs substantial training costs. In this paper, we propose a novel concept-based curriculum masking (CCM) method to efficiently pre-train a language model. CCM has two key differences from existing curriculum learning approaches to effectively reflect the nature of MLM. First, we introduce a carefully-designed linguistic difficulty criterion that evaluates the MLM difficulty of each token. Second, we construct a curriculum that gradually masks words related to the previously masked words by retrieving a knowledge graph. Experimental results show that CCM significantly improves pre-training efficiency. Specifically, the model trained with CCM shows comparative performance with the original BERT on the General Language Understanding Evaluation benchmark at half of the training cost. Code is available at \href{https://github.com/KoreaMGLEE/Concept-based-curriculum-masking}{https://github.com/KoreaMGLEE/Concept-based-curriculum-masking}.

\end{abstract}

\section{Introduction}
Self-supervised pre-training has achieved considerable performance improvements in various natural language processing (NLP) tasks \cite{Devlin-NAACL:2019, Yang-NeurIPS:2019, Zhang-ACL:2019, Lewis-ACL:2020, Clark-ICLR:2020}. Masked language modeling (MLM) \cite{Devlin-NAACL:2019}, where the model predicts original tokens of a masked subset of the text using the unmasked subset as clues, has contributed significantly to these improvements. However, MLM typically requires a high number of compute operations, resulting in unrealistically large training costs \cite{Clark-ICLR:2020}.
Therefore, the consideration on the pre-training costs has become an increasingly important issue \cite{Jiang-NeurIPS:2020, Narayanan:2021}.

\begin{table}
\begin{center}
\setlength\tabcolsep{3.3pt}
\begin{tabular}{lcccc}
\hline
Methods        & CoLA &  MRPC & SST  & Avg.  \\ \hline
No curriculum  & 44.9 & 85.4 & 89.6  & 73.3 \\\hline 
Rarity & 45.5 & 85.1 & 89.4  & 73.4 \\ 
Length & 41.6 & 86.3 & 89.4 & 72.4 \\
Masking ratio & 43.2 & \textbf{87.0} & 89.7 & 73.3 \\ \hline
CCM (ours)  & \textbf{48.0} & 86.7 & \textbf{90.9}  & \textbf{75.2} \\\hline
\end{tabular}
\end{center}
\caption{Comparison of different CL methods with BERT\textsubscript{Medium}. Rarity denotes a curriculum that first learns sentences composed with frequent words. Length denotes a curriculum that incrementally increases the sequence length from 64 to 512. Masking ratio denotes a curriculum that increases the masking ratio linearly from 0.1 to 0.15.}
\label{tab:my-table}
\end{table}

Recent NLP studies have shown that curriculum learning (CL), presenting examples in an easy-to-difficult order rather than presenting them randomly, can accelerate the model convergence and improve the generalization performance \cite{Zhang-CoRR:2018, Tay-ACL:2019, Zhan-AAAI:2021}. There mainly exist two criteria for assessing the difficulty of examples, 1) model-based criteria \cite{Zhou-ACL:2020, Xu-ACL:2020} and 2) linguistic-inspired criteria \cite{Sachan-ACL:2016, Tay-ACL:2019, Nagatsuka-RANLP:2021, Campos-arXiv:2021}. Model-based criteria measure the difficulty of each example using task-specific models. 
However, these criteria are unsuitable for reducing the computation cost of pre-training, given that they require calculating the loss of every example in a large pre-training corpus using language models. In contrast, linguistic-inspired criteria can efficiently assess the difficulty of examples based on prior knowledge and rules. Therefore, we adopt CL with linguistic difficulty criteria into MLM to improve the efficiency of pre-training.

However, we argue that existing linguistics-inspired criteria, such as length, rarity, and masking ratio of a sequence, do not effectively reflect the nature of MLM, as verified empirically in Table 1. The difficulty associated with MLM is significantly affected by the choice of tokens to be masked in the given sequence, rather than by the given sequence itself. For example, given "The man is a Stanford \textsf{<mask>} student", we can easily predict that the masked token would be \textsf{University}, whereas given "The man is a \textsf{<mask>} University student", it would be relatively difficult to predict the original token due to the insufficient clues in the context. Then, how can we measure the MLM difficulty? MLM can be viewed as predicting masked tokens based on other contextual tokens related to masked tokens. Therefore, if a word is related to many other words and phrases, it is likely that it has several clues in the context that make MLM easier.

In this paper, we propose a novel concept-based curriculum masking (CCM) for improving pre-training efficiency by considering the nature of MLM. We consider words and phrases that are related to several other concepts as easy ones and define them as the initial concepts to be masked first. To identify them, we utilize millions of syntactic and semantic relationships between words or phrases, referring as concepts, within a large-scale knowledge graph, ConceptNet \cite{Speer-AAAI:2017}. First, we measure the number of related concepts for each concept in ConceptNet and construct the set of concepts with the highest number, which will be masked during the first stage of our curriculum. Then, we gradually mask concepts related to the previously masked concepts during the consecutive stages, inspired by human language acquisition, in which simple concepts are learned first and more complex concepts are gradually learned \cite{Anglin-Child:1978, Horton-Child:1980}. 

To verify the effectiveness of the proposed curriculum, we conduct experiments on the General Language Understanding Evaluation (GLUE) benchmark \cite{Wang-ICLR:2019}. Experimental results show that CCM significantly improves the pre-training efficiency of MLM. Particularly, CCM only requires 50\% of the computational cost in achieving the comparable GLUE score with MLM, using the same BERT\textsubscript{Base} architecture. In addition, by training on commensurate computational cost as if MLM, CCM outperforms MLM by 1.9 points on the GLUE average score. The contributions of our study are as follows:

\begin{itemize}
  \item We investigate and demonstrate the effectiveness of CL for pre-training. To the best of our knowledge, our work is one of the few that introduces CL to MLM. 
  \item We propose a novel curriculum masking framework that progressively masks concepts based on our relation-based difficulty criteria.
  \item We demonstrate that our framework significantly improves the pre-training efficiency of MLM through extensive experiments on the GLUE benchmark.
\end{itemize}

\begin{figure*}[!t]

\centering
\includegraphics[width=\textwidth]{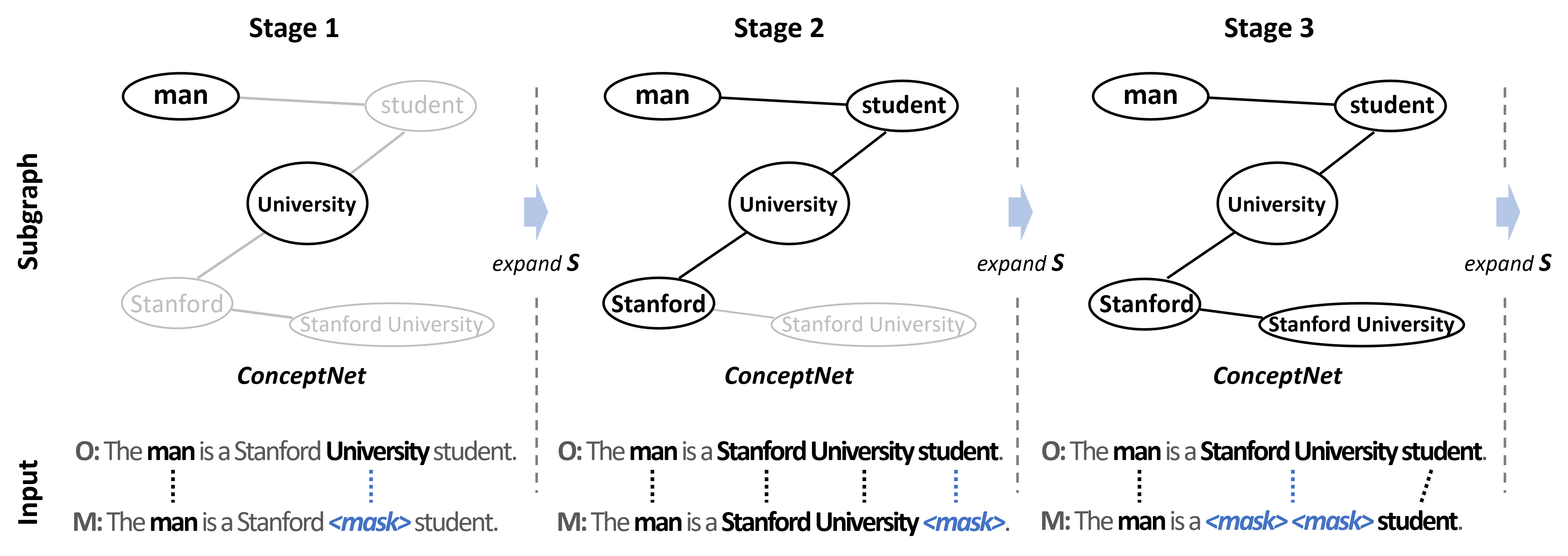}
\caption{Illustration of CCM. In the first stage, CCM masks the initial concepts. Then it progressively includes the concepts within $k$-hops from the concepts used in the previous stages. Note that, "S" represents a set of concepts to be masked, "O" represents the original token sequence, and "M" represents a masked token sequence.}

\label{fig:label1}
\end{figure*}

\section{Related Work}
\subsection{Self-supervised Pre-training}
Self-supervised pre-training has been employed to learn universal language representations from large corpora \cite{Collobert-jMach:2011, Pennington-EMNLP:2014}. Recently, BERT \cite{Devlin-NAACL:2019} has achieved tremendous success in various NLP tasks by learning bidirectional representations via MLM. Variants of BERT have been proposed to further improve the performance in NLP tasks. \citet{Yang-NeurIPS:2019} have used an input sequence autoregressively and randomly generated to alleviate the mismatch caused by the masked tokens that appear only during pre-training. \citet{Zhang-ACL:2019} have incorporated external knowledge graphs into the language model to introduce structural knowledge. \citet{Sun-CoRR:2019} and \citet{Joshi-TACL:2020} have masked contiguous tokens to improve the span representation. \citet{Lewis-ACL:2020} have replaced multiple tokens with one mask token for noise flexibility. \citet{Levine-ICLR:2021} have masked highly collocating n-grams in the corpus for preventing the model to use shallow local signals. Although these studies have resulted in significant performance improvements in NLP tasks, they still incur tremendous computing costs \cite{Qiu-CoRR:2020}. To the best of our knowledge, there are only limited studies that have explored the improvement of MLM pre-training efficiency.

\subsection{Curriculum Learning}
In the domain of machine learning, CL is a training strategy that gradually learns the complex examples after learning the easy ones, instead of learning all data simultaneously, which has been firstly proposed by \cite{Bengio-ICML:2009}. The concept of CL can be traced back to \citet{elman-Cognition:1993} that has attempted to train machines from easy to difficult tasks. \citet{Krueger-Cognition:2009} have tried to divide complex tasks into easy sub-tasks, and then trained machines on these sub-tasks.

Recent studies have shown that a curriculum-based approach can improve the convergence speed and generalization performance of NLP systems \cite{Sachan-ACL:2016, Zhang-CoRR:2018, Tay-ACL:2019, Zhou-ACL:2020, Zhan-AAAI:2021}. The identification of easy samples in a given training dataset is important in applying CL to a target task \cite{Kumar-NeurIPS:2010}. There are two main strategies for assessing the data difficulty; 1) using human prior knowledge regarding specific tasks \cite{Zhang-CoRR:2018, Tay-ACL:2019, Zhan-AAAI:2021}, and 2) using trained models to measure the difficulty of examples \cite{Zhou-ACL:2020, Xu-ACL:2020}. However, existing difficulty criteria do not work well with CL for MLM, as they have overlooked that the MLM difficulty of examples can be changed according to which tokens are masked. To better assess the difficulty of MLM, we propose a novel curriculum that intervenes in the masking process based on the relation-based difficulty criteria.

\section{Concept-based Curriculum Masking}
Our curriculum framework is inspired by the human language acquisition. People learn simple concepts (e.g., car) first and then gradually learn more complex concepts (e.g., self-driving car). By leveraging the related basic concepts, the gradual learning style enables humans to learn abstract concepts easily \cite{Anglin-Child:1978, Horton-Child:1980}. To mimic this, we construct a multi-stage curriculum by gradually adding concepts relating to the initial concepts using a knowledge graph. To this end, our CCM framework consists of three processes: initial concept selection, curriculum construction, and pre-training. 

In the following sections, we denote a knowledge graph as $G=(N, E)$, where $N$ and $E$ are sets of nodes and edges, respectively. We define a word or phrase corresponding to a node as a concept $c \in N$. 
We denote $S_{i}$ as the set of concepts to be masked in the $i$-th stage. The final curriculum includes multiple stages $\left \{ {S}_{1}... {S}_{K} \right \}$.

\subsection{Initial Concept Selection}
A key for CL is to learn easy examples first. Considering that MLM is the task for predicting a masked subset of a text using the surrounding context, we suppose concepts related to many others are easy concepts. Thus, we construct a set of initial concepts for masking by selecting concepts that have the highest degree of connection in the knowledge graph. To select the initial concepts, we first rank each concept $c \in N$ according to the degree of connection in the knowledge graph $G$. In addition, we exclude concepts that appear less than 100k times in the pre-training corpus, since frequent concepts are more influential in learning their related concepts than rare concepts.\footnote{We have observed that rare concepts connected to many concepts are mostly medical or scientific jargons, which are intuitively considered to be complex (e.g., heraldry, carbohydrate).} Then, given the pre-defined number of initial concepts $M$, the top $M$ concepts are selected. 

\subsection{Curriculum Construction}
In this section, we describe how to arrange the learning stages based on the knowledge graph $G$. The set of concepts to be masked in the $i$-th stage, $S_i$, is constructed following the principle
\begin{quote}
    \textit{\textbf{$S_{i}$ is constructed by progressively expanding $S_{i-1}$ to concepts that are related to concepts in $S_{i-1}$.}}
\end{quote}
This facilitates to utilize previously learned concepts as clues for modeling new related concepts like human language acquisition. We identify related concepts using the relationships in the knowledge graph. Intuitively, the more closely connected the two concepts are in the knowledge graph, the more related they are. For example, as shown in Figure 1, given a concept \textsf{University}, \textsf{student} and \textsf{Stanford} (1-hop distance) are more related than \textsf{man} (2-hop distance). Consequently, \textsf{student} or \textsf{Stanford} may be utilized as a stronger predictive clue than \textsf{man}, such as in the sentence "The man is a Stanford \textsf{<mask>} student" where \textsf{University} is masked. 

Based on the principle, we gradually mask concepts connected to previously learned concepts within $k$-hops in the knowledge graph throughout curriculum stages as follows:
\begin{equation}
     {S}_{i} = {S}_{i-1} \cup {N_{k}({S}_{i-1})}
\end{equation}
Here, $N_{k}({S}_{i-1})$ denotes the set of concepts, each concept of which is within $k$-hops from $s \in S_{i-1}$. 

\subsection{Pre-training}
To introduce concept-based curriculum masking into pre-training language models, we first search for concepts included in $S_{i}$ for each $i$-th stage from the token sequences in the pre-training corpus and then mask identified concepts for MLM training.

\paragraph{Concept search} We first compile a lexicon of concepts. Specifically, from the knowledge graph, we extract the concepts of less than 5 words that occur over 10 times in the pre-training corpus. Then we search for the extracted concepts from the token sequences in the pre-training corpus by string match. The cost of this process is negligible compared to MLM as it is conducted only once during pre-processing.\footnote{In our experiment, it takes about 46 minutes to search for concepts in the pre-training corpus with two Intel(R) Xeon(R) Silver 4210R CPUs.}

\begin{algorithm}[t]
\centering
\caption{Curriculum Masking Scheme}\label{alg:cap}
\begin{flushleft}
\textbf{Input:} Concept-based curriculum $\left \{ {S}_{1}... {S}_{K} \right \}$, language model parameters $\theta$, input data $D$, maximum training step $\tau$, dynamic masking probability ${p}_{d}$ 
\begin{algorithmic}[1]

\For{token sequence $T$ in $D$}
\State Search concepts $c $ in $T$.
\State Ordered concepts according to $\left \{ {S}_{1}... {S}_{K} \right \}$
\EndFor

\State Initialize $\theta$

\While{step < $\tau$}
\For{stage $i = 1, ..., K$}

\State Generate examples $e$ by masking $c$ $(c \in S_{i}$ and $c \in T)$ based on a rate of ${p}_{d}$. 
\State Training the model on $e$.

\EndFor
\EndWhile
\State \Return Trained model parameters $\theta$

\end{algorithmic}
\end{flushleft}
\end{algorithm}

\begin{figure*}[ht]
\begin{subfigure}{.33\textwidth}
  \centering
  \includegraphics[width=0.9\linewidth]{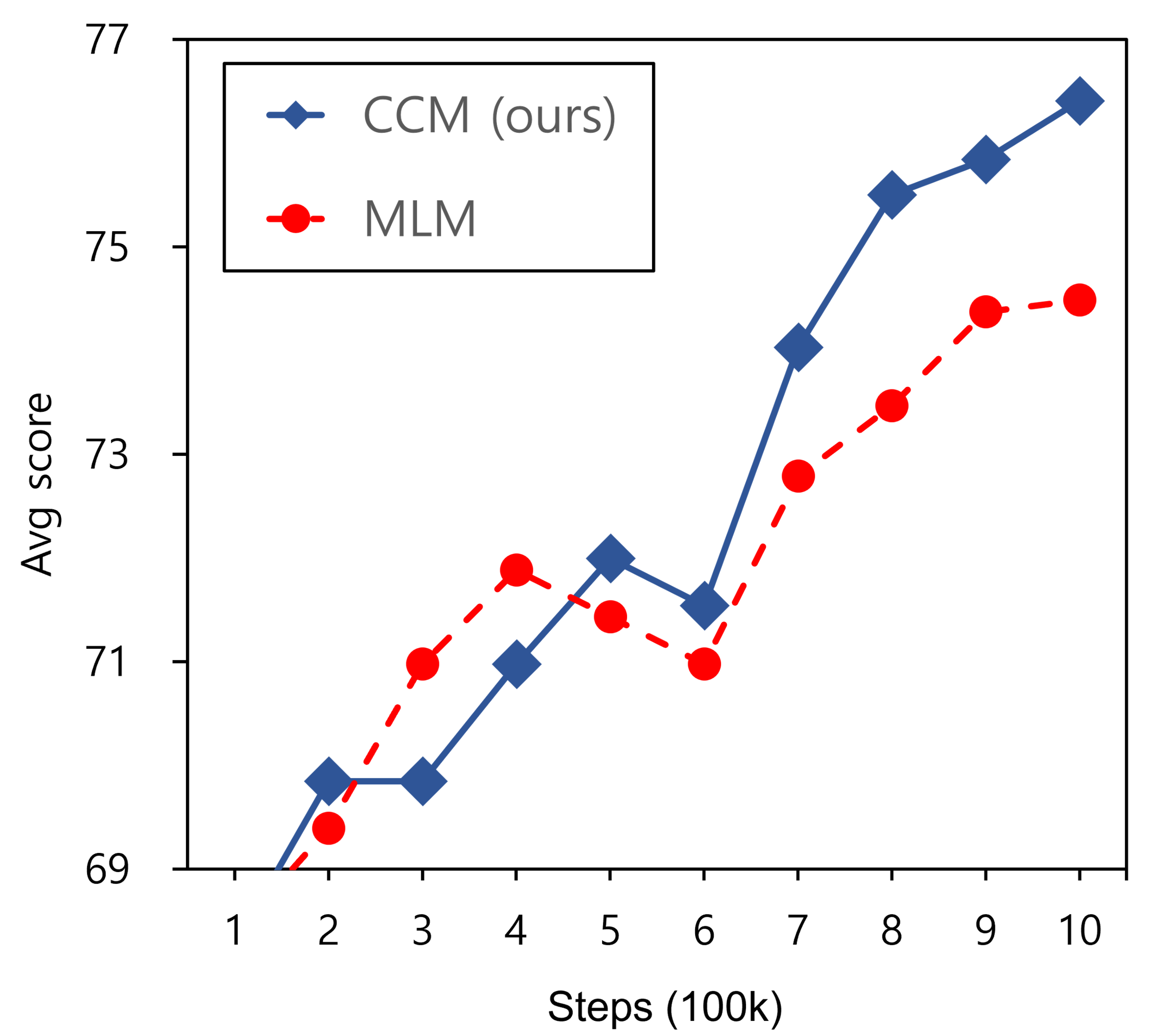}  
  \ \ \ \ \ \caption{Small-sized model}
  \label{fig:sub-first}
\end{subfigure}
\begin{subfigure}{.33\textwidth}
  \centering
  \includegraphics[width=0.9\linewidth]{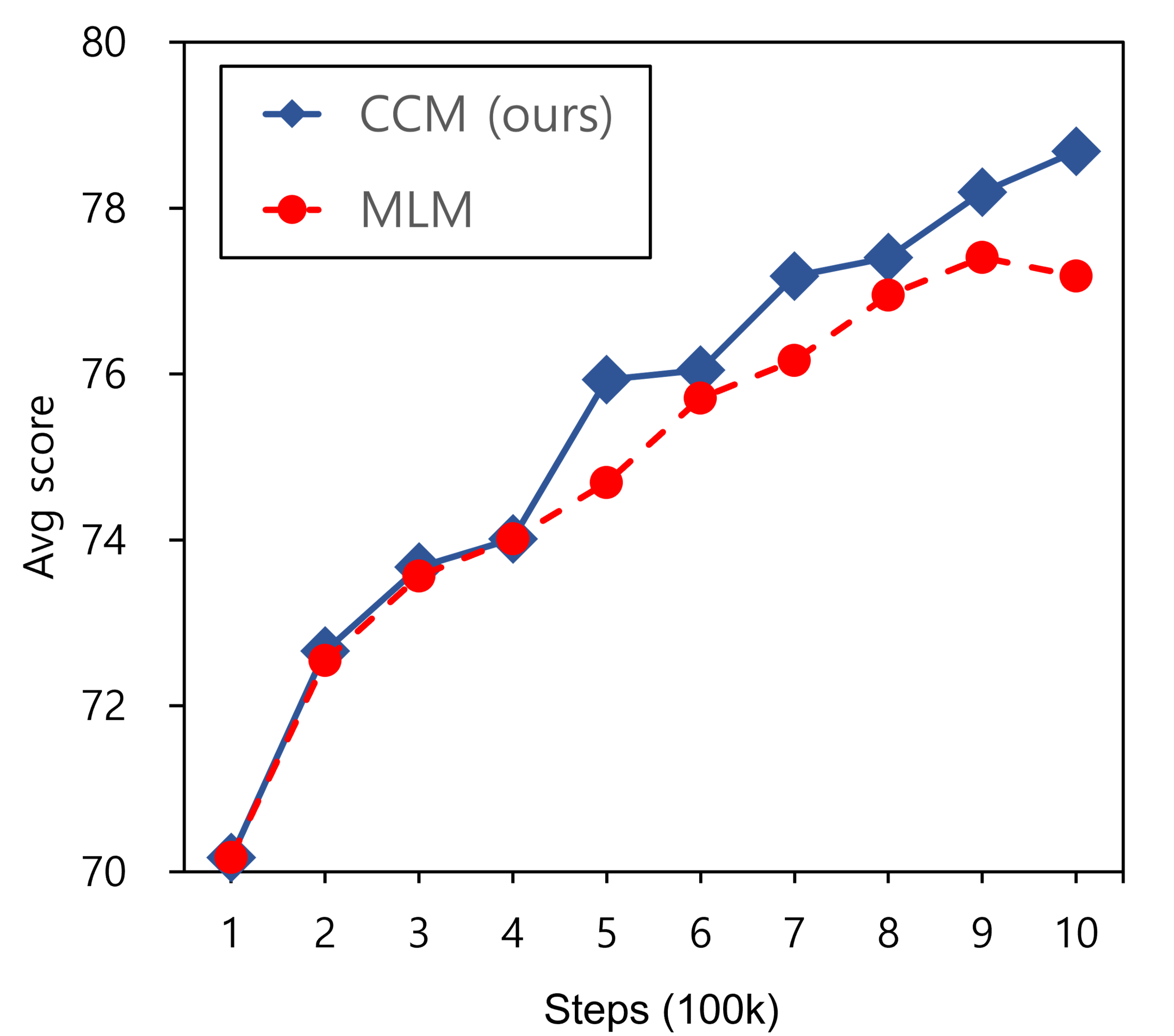}  
  \ \ \ \ \caption{Medium-sized model}
  \label{fig:sub-first}
\end{subfigure}
\begin{subfigure}{.33\textwidth}
  \centering
  \includegraphics[width=0.9\linewidth]{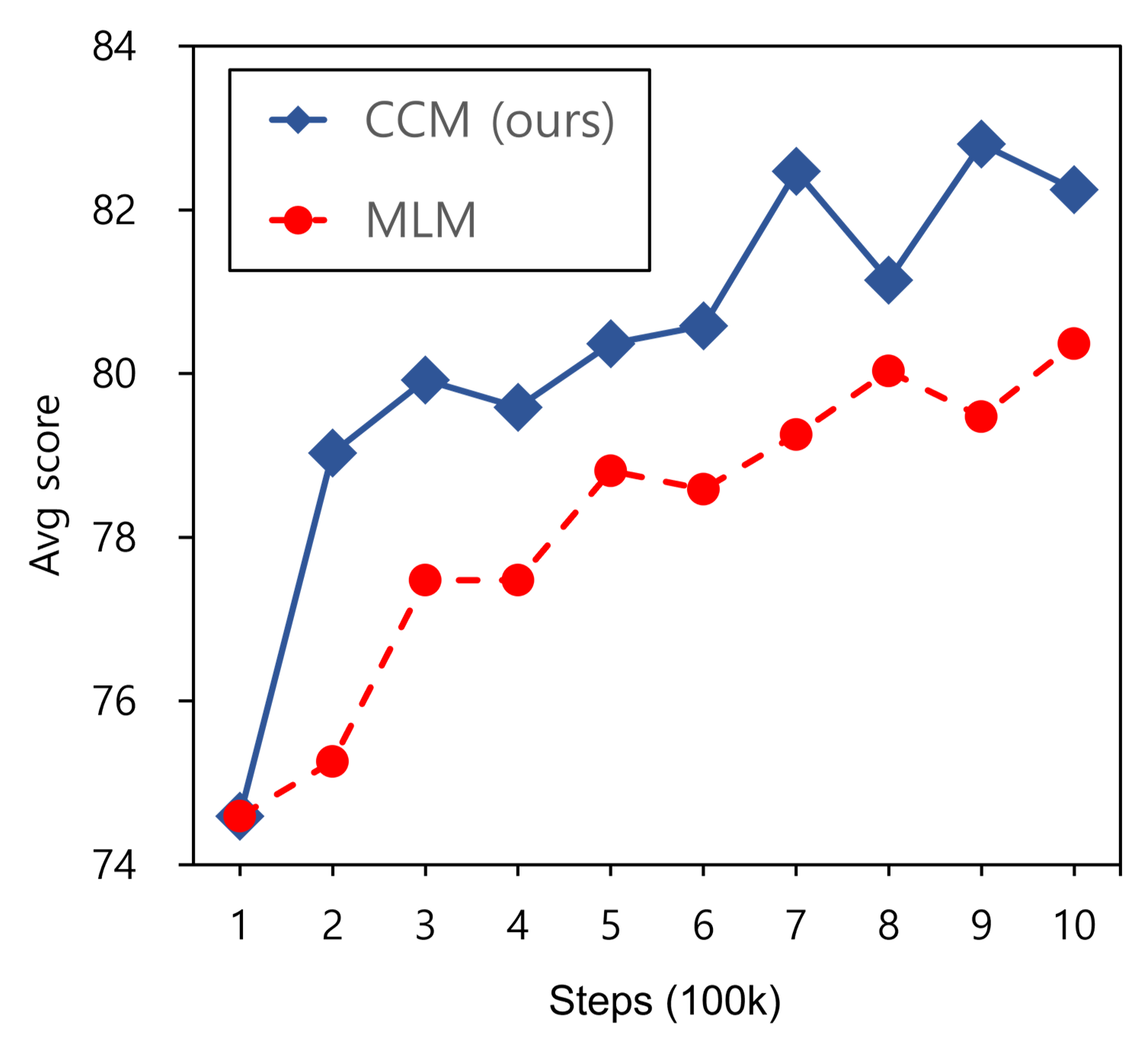}  
  \ \ \ \ \ \caption{Base-sized model}
  \label{fig:sub-second}
\end{subfigure}
\caption{Comparison of MLM and CCM on different-sized BERT models. The reported results are average scores on the GLUE benchmark with respect to steps.}
\label{fig:fig}
\end{figure*}

\begin{table*}
\begin{center}
\setlength\tabcolsep{4pt}
\begin{tabular}{llccccccccccc}
\hline

Model   & Method & Params  & CoLA & SST  & MRPC & STS & RTE & MNLI & QQP & QNLI & Avg.  \\ \hline
\multirow{2}{*}{BERT\textsubscript{Small}}  & MLM & 14M & 38.0 & 88.7 & 82.8 & 82.0
 & 59.2 & 76.8 & 88.4 & 85.8 & 75.2 \\
&  CCM (ours) & 14M & \textbf{42.8} & \textbf{89.1} & \textbf{84.1} & \textbf{83.3} & \textbf{61.3} & \textbf{77.5} & \textbf{88.6} & \textbf{86.3} & \textbf{76.6} \\ \hline

\multirow{2}{*}{BERT\textsubscript{Medium}} & MLM & 26M &44.9 & 89.6 & 85.4 & 82.7 & 60.3 & 78.9 & \textbf{89.4} & \textbf{87.6} & 77.4 \\
&  CCM (ours) & 26M & \textbf{48.0} & \textbf{90.9} & \textbf{86.7} & \textbf{83.6} & \textbf{61.4} & \textbf{80.0} & 89.2 & \textbf{87.6} & \textbf{78.4} \\\hline

\multirow{2}{*}{BERT\textsubscript{Base}} &  MLM & 110M & 49.7 & 90.8 & 87.8 & 85.4 & \textbf{67.8} & 81.7 & 90.4 & 89.5 & 80.4 \\
&  CCM (ours) & 110M & \textbf{60.3} & \textbf{93.1} & \textbf{88.3} & \textbf{85.5} & 65.0 & \textbf{84.1} & \textbf{91.0} & \textbf{91.4} & \textbf{82.3} \\ \hline

\end{tabular}
\end{center}
\caption{Results of small, medium, and large-sized models on the development sets of GLUE.}
\label{tab:my-table}
\end{table*}

\paragraph{Concept masking} After searching for the concepts, we mask the token sequences following our curriculum. We introduce whole concept masking (WCM) that masks all the tokens consisting of a single concept simultaneously. For example, if we mask the concept \textsf{Stanford}, all the tokens \textsf{Stan} and \textsf{\#\#ford} will be masked together. Following the masking strategy in \cite{Devlin-NAACL:2019}, 80\% of the total masked concepts are replaced into mask tokens, and 10\% into random tokens, and the rest are not replaced. 

The number of identified concepts changes significantly for each stage and sentence. Hence, the static masking probability often leads to either too little or too much masking. Therefore, we dynamically calculate a masking probability $p_{d}$ to mask approximately 15\% of the total tokens for each sequence.

In addition, even if a concept includes other concepts in a given input, all the concepts are treated independently. For example, if two concepts \textsf{Stanford University} and \textsf{Stanford} appear in a sequence, each concept is masked independently. The entire CCM process is presented in Algorithm 1.

\section{Experiment}
We examine the efficacy of CCM using the BERT architectures \cite{Devlin-NAACL:2019}. We measure the performance of the pre-trained models on the GLUE benchmark \cite{Wang-ICLR:2019}.

\subsection{Experimental Settings}
In the experiments, we denote the token embedding size as $E$, hidden dimension size as $H$, number of layers as $L$, intermediate layer size of the feed-forward module as $F$, and the number of self-attention heads as $A$, respectively. We report experimental results on three model sizes: Small ($E$ = 128, $H$ = 256, $L$= 12, $F$ = 1024, $A$ = 4, 14M parameters), Medium ($E$ = 128, $H$ = 384, $L$ = 12, $F$ = 1536, $A$ = 8, 26M parameters), and  Base ($E$ = 768, $H$ = 768, $L$ = 12, $F$ = 3072, $A$ = 12, 110M parameters). We conduct experiments with four RTX A5000 GPUs.

\paragraph{Pre-training.} Since the performance of pre-trained language models heavily depends on the corpus size, we manually pre-train all the BERT models using CCM and MLM on the BERT pre-training corpus released in HuggingFace Datasets\footnote{https://huggingface.co/datasets}, including BooksCorpus \cite{Zhu-ICCV:2015} and the English Wikipedia to ensure a fair comparison. 
We pre-train small and medium-sized models for 1M steps with a batch size of 128, a sequence length of 128, and a maximum learning rate of 5e-4. Furthermore, we pre-train base-sized models for 1M steps with a batch size of 256, a sequence length of 128, and a maximum learning rate of 1e-4. We use Adam as optimizer with $\beta_{1} = 0.9, \beta_{2} =0.999,$ and L2 weight decay of 0.01. For pre-training, after a warmup of 10k steps, we used a linear learning rate decay. It is noteworthy that during CCM, we warmup for 100k steps using MLM and then train 100k steps for each stage. When the final stage ends, the model returns to the MLM stage and repeats the curriculum for the remaining steps. In these experiments, we use a four-stage curriculum, where the final stage can mask all concepts and all words that do not comprise a concept. We pre-train 3 randomly initialized models and use the model with the lowest validation MLM loss.

\paragraph{Evaluation.} We evaluate our pre-trained models on the GLUE benchmark. The GLUE benchmark consists of eight datasets for the evaluation of natural language understanding systems: RTE \cite{Giampiccolo-ACL-PASCAL:2007} and MNLI \cite{Williams-Mach:1992} cover textual entailment; QNLI \cite{Rajpurkar-EMNLP:2016} covers question-answer entailment; MRPC \cite{Dolan-IWP@IJCNLP:2005} covers paraphrase; QQP covers question paraphrase; STS \cite{Cer-SemEval:2017} covers textual similarity; SST \cite{Socher-EMNLP:2013} covers sentiment; and CoLA \cite{Warstadt-Trans:2019} covers linguistic acceptability. We use the Mathew correlation for CoLA, Spearman correlation for STS, and accuracy for the rest of the tasks as evaluation metrics. We further report an average score over the eight tasks. 

We fine-tune the pre-trained models on CoLA, MRPC, SST, QQP, QNLI, and MNLI for three epochs while STS and RTE for 10 epochs. For each task, we select the best fine-tuned learning rates (among the 5e-5, 4e-5, 3e-5, and 2e-5 following the setting in \citet{Devlin-NAACL:2019}) on the development sets. In addition, we run five random restarts and report the median score. For each random restart, we use the same checkpoint but with different data shuffling and classifier initialization methods. Note that we conduct all experiments using a single model, not using an ensemble.

\begin{table}
\begin{center}
\begin{tabular}{llc}
\hline
Model                          & Method  & GLUE  \\ \hline
\multirow{3}{*}{BERT\textsubscript{Medium}}  & CCM (ours)        & \textbf{78.4}\\ 
                             & \ \ w/o CL                   & 76.8 \\ 
                             & \ \ w/o CL, WCM               & 77.4  \\ \hline
\end{tabular}
\end{center}
\caption{Ablation studies of CCM using medium-sized BERT architecture. Here, "WCM" represents whole concept masking and "CL" represents curriculum learning.}
\label{tab:my-table}
\end{table}

\subsection{GLUE Results}
As aforementioned, the goal of CCM is to accelerate the pre-training convergence. By examining the performance of models on the GLUE benchmark after every 100k steps, we demonstrate that CCM accelerates model convergence speed. 

Figure 2 shows the learning curves of different size models on the GLUE benchmark. We observe that CCM allows all models to achieve baseline performance with significantly less computation. In particular, CCM outperforms MLM at 50\% of the computational cost with base-sized models. In addition, we can observe that CCM is more effective with a larger model. With a small-sized model, CCM shows slightly lower performance during the initial pre-training steps. We speculate that concept-wise masking may be too hard for a small-sized model, resulting in a degradation of convergence speed. Nevertheless, after training with enough iterations, CCM can achieve the GLUE scores of MLM with fewer training steps.

Table 2 shows the performance of the models pre-trained for 1M steps on each GLUE task. We can observe that CCM allows all models to outperform their baselines on all tasks, except for BERT\textsubscript{Base} on RTE. Specifically, CCM-applied BERT\textsubscript{Small} and BERT\textsubscript{Medium} outperform their baselines by 1.4 points and 1.0 points on the GLUE average score, respectively, when fully trained. CCM-applied BERT\textsubscript{Base} achieves the best performance improvement on GLUE average score. Although CCM-applied BERT\textsubscript{Base} shows worse performance on RTE, possibly due to the significantly different concept distribution compared with the pre-training corpus, we have observed CCM-applied BERT\textsubscript{Base} finally outperforms the baseline performance with slightly more training steps.

\section{Analysis}
\subsection{Ablation Study}
To investigate the contribution of each component in CCM, we conduct ablation studies for whole concept masking (WCM) and CL. The ablation study is conducted with medium-sized models and GLUE scores are reported. The GLUE score is the average score of all eight tasks. As shown in Table 3, we can observe that CCM achieves the best GLUE scores when using both CL and WCM while having the worst score with the setting without CL. These results indicate that CL greatly contributes to the improvements in pre-training efficiency.

\begin{table}
\begin{center}
\setlength\tabcolsep{3.3pt}
\begin{tabular}{lcccc}
\hline
Methods        & CoLA &  MRPC & SST  & Avg.  \\ \hline
No curriculum  & 44.9 & 85.4 & 89.6  & 73.3 \\\hline 
Rarity & 45.5 & 85.1 & 89.4  & 73.4 \\ 
Length & 41.6 & 86.3 & 89.4 & 72.4 \\
Reverse   & 29.0 & 83.6 & 87.6  & 66.7 \\ 
Masking ratio & 43.2 & \textbf{87.0} & 89.7 & 73.3 \\ 
Teacher review & \textbf{48.0} & 85.4 & 89.9  & 74.4 \\ \hline
CCM (ours)  & \textbf{48.0} & 86.7 & \textbf{90.9}  & \textbf{75.2} \\\hline
\end{tabular}
\end{center}
\caption{Comparison of different curriculum designs with BERT\textsubscript{Medium}.}
\label{tab:my-table}
\end{table}

\subsection{Effect of the Curriculum}
To demonstrate our curriculum design choice, we compare it with the non-curriculum and other curriculum-based approaches on BERT\textsubscript{Medium}. For comparison with other curricula, we adopt rarity, length, reverse, masking ratio, and teacher review as baselines. All curriculum-based approaches use the same curriculum configuration (e.g., the number of stages) unless mentioned otherwise. 

\paragraph{Rarity.} We adopt the rarity of concepts in the training corpus as a difficulty metric, similar to those used in \cite{Tay-ACL:2019}. We initially train the model with the most frequent concepts in the training corpus, and then progressively add the less frequent ones. 

\paragraph{Length.} We first train the model on short token sequences and then progressively train on longer sequences. Following \cite{Campos-arXiv:2021}, we initially use a sequence length of 64, then gradually increase the length to 128, 256, and 512 at the end of each epoch. 

\paragraph{Reverse.}  For comparison with a hard-to-easy curriculum, we train the model with the reverse order of curriculum in CCM. Specifically, MLM is used to warmup the hard-to-easy model for 100k steps, subsequently, we train the model from stage 3 to stage 1.

\paragraph{Teacher review.} For teacher review, the BERT\textsubscript{Base} model is used as the teacher. We use the MLM loss from the pre-trained teacher as the difficulty score \cite{Xu-ACL:2020} and distribute examples according to the measured difficulty.

\paragraph{Masking ratio.} For the masking ratio curriculum, we only mask 10\% of the first full sequence. Subsequently, we increase the masking ratio linearly to 15\% of tokens when 1M is reached.

\paragraph{Results.} As shown in Table 4, all curriculum-based approaches except the reverse curriculum improve generalization performance on various tasks compared to non-curriculum. The hard-to-easy curriculum shows a significant performance degradation. A possible reason could be that concepts added in the last step are too difficult for language models to learn without prior learning any relevant concepts, leading to the degradation of convergence speed.
Finally, our CCM outperforms all other approaches in the experiment on the GLUE tasks. These results indicate that our curriculum for progressively learning relevant concepts is effective to improve pre-training efficiency.

\begin{figure}[]
\includegraphics[width=0.5\textwidth]{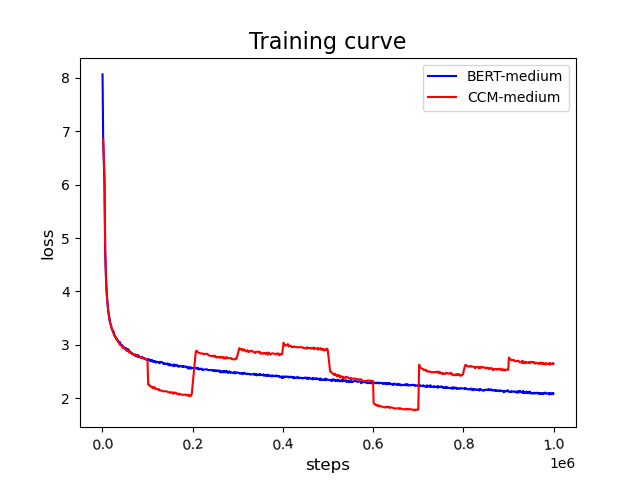}
\caption{Training curve of the medium-sized models on pre-training. We report the training loss at every hundred steps.}
\label{fig:label1}
\end{figure}

\begin{table}[t]
\begin{center}
\begin{tabular}{ll}
\hline
           & Loss mean \\ \hline

Warmup    & 1.95\textpm0.61 \\
Stage 1    & \textbf{1.89\textpm0.53}  \\
Stage 2    & 2.50\textpm0.95 \\ 
Stage 3    & 2.72\textpm1.14 \\
Stage 4    & 2.42\textpm0.93\\\hline
\end{tabular}
\end{center}
\caption{Inference loss of MLM-applied BERT\textsubscript{Base} for each curriculum stage.}
\label{tab:my-table}
\end{table}

\subsection{Difficulty of Curriculum Stages}
Most CL studies argue that the generalization performance and convergence speed are improved when training models in an easy-to-difficult order. To examine whether our curriculum arranges examples in an easy-to-difficult order as in \citet{Bengio-ICML:2009}, we measure the difficulty of examples in each stage of our curriculum. For measuring the difficulty of examples, we report training loss of the CCM-applied BERT\textsubscript{Medium} model at every hundred steps. In addition, we evaluate a mean loss of 3,000 examples for each stage using pre-trained BERT\textsubscript{Base}, consistent with \citet{Xu-ACL:2020}. 

Figure 3 shows a training curve of CCM-applied BERT\textsubscript{Medium}. We observe a trend of increasing losses from stage 1 to stage 4 during training, which validates that our curriculum effectively trains the model in an easy-to-difficult order. In addition, Table 5 shows the inference MLM loss of pre-trained BERT. In Table 5, we also observe a similar tendency to Figure 3, except for stage 4. We expect that this is due to the difference in the masking distribution on which the model is trained in this experiment. 

\subsection{Analysis of Initial Concept}
We analyze the effect of the number of initial concepts and the selection criteria in medium-sized models to gain a better understanding of the initial concept. The detailed results are presented in Table 6, Table 7, and Figure 4. 

\paragraph{Effect of the number of concepts.}
We first investigate the effect of the number of initial concepts by changing the number of the first stage concepts from 1k to 10k and analyze the impact on the performance. As shown in Table 6, the highest and lowest performances are obtained when the number of concepts is 3k, and 1k, respectively. Specifically, in the case of 1k, CCM does not have any improvement over the MLM baseline in SST and MRPC tasks. These results imply that some enough number of initial concepts should be learned for our method to be effective (e.g., 3k in our experimental setting).

\begin{table}
\begin{center}
\begin{tabular}{ccccc}
\hline
\#concepts & CoLA & MRPC & SST & Avg.  \\ \hline
MLM & 44.9 & 85.4 & 89.6 & 73.3 \\ \hline
1k & 47.6 & 85.1 & 88.9 & 73.8 \\ 
3k & \textbf{48.0} & \textbf{86.7} & \textbf{90.9} & \textbf{75.2}\\
5k & 46.8 & 86.2 & 90.7 & 74.5 \\
10k & 46.6 & 86.1 & 90.7 & 74.4 \\\hline

\end{tabular}
\end{center}
\caption{Comparison of different number of initial concepts.}
\label{tab:my-table}
\end{table}

\begin{table}
\begin{center}
\setlength\tabcolsep{4.5pt}
\begin{tabular}{ccccc}
\hline
Initial concepts                           & CoLA & MRPC & SST & Avg.  \\ \hline

 HF        & 44.3 & 85.9 & 89.1 & 73.1  \\
 RC          & 47.7 & 86.1 & 89.3 & 74.3  \\
 HF, RC  & \textbf{48.0} & \textbf{86.7} & \textbf{90.9} & \textbf{75.2} \\ \hline

\end{tabular}
\end{center}
\caption{Comparison of different criteria for initial concepts. Here, "HF" represents high frequency and "RC" represents the number of related concepts.}
\label{tab:my-table}
\end{table}

\begin{figure}[t]
\includegraphics[width=0.5\textwidth]{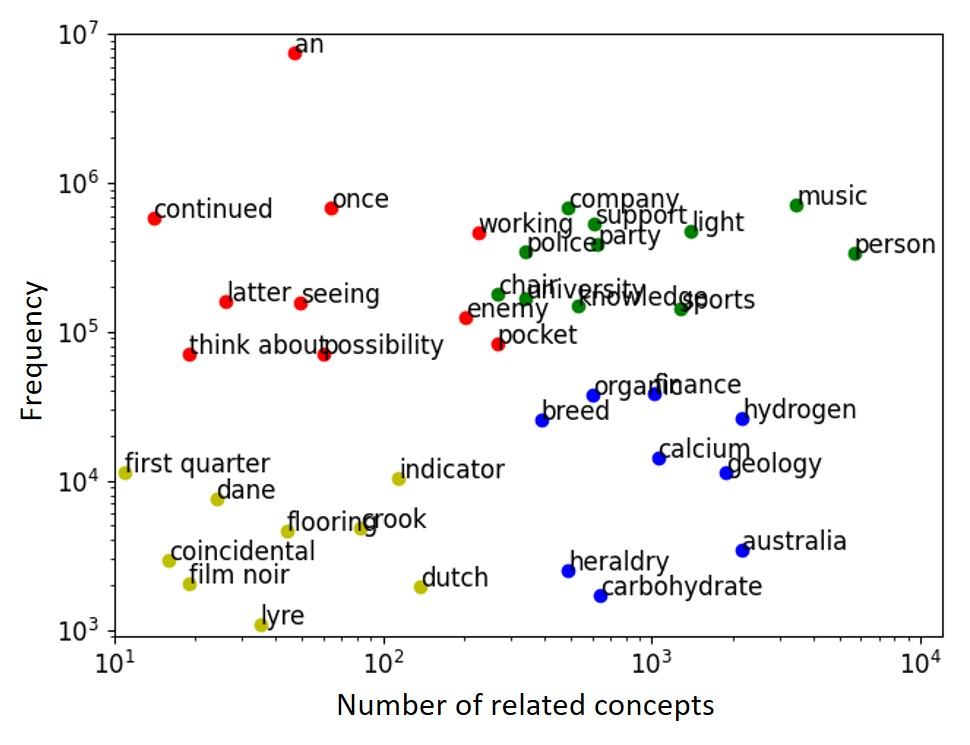}
\caption{A plotted graph demonstrating some concrete examples of concepts according to the balance between the "frequency" and the "number of related concepts". Each color represents a group of classes. For example, red refers to concepts with a high frequency but a low number of related concepts.}

\label{fig:label1}
\end{figure}

\paragraph{Effect of selection criteria.}
To investigate the influence of the criteria for selecting the easiest concepts, we compare with the criteria of selecting the easiest concepts from 1) many related concepts, 2) high-frequency concepts, and 3) high frequency-many related concepts. As shown in Table 7, our criterion for high frequency-many related concepts performs the best among other criteria. In addition, Figure 4 shows examples of concepts classified by the above criteria. It shows that when a concept is selected only by frequency, stopwords such as \textsf{an} or \textsf{the} would be included, and when selecting only by the number of related concepts, many academic terms such as \textsf{carbohydrate} would be included in the initial concepts. These results indicate that concepts that satisfy both criteria simultaneously are easier to learn and helpful in learning other concepts later.

\begin{table}
\begin{center}
\begin{tabular}{lcccc}
\hline
Methods        & MRPC & SST & STS & Avg.  \\ \hline

CCM\textsubscript{1-hop}  & 85.8 & 89.9 & \textbf{83.6} & 86.4\\
CCM\textsubscript{2-hop}  & \textbf{86.7} & \textbf{90.9} & \textbf{83.6} & \textbf{87.0} \\
CCM\textsubscript{3-hop}  & 84.9 & 89.2 & 82.8 & 85.6 \\ 
CCM\textsubscript{4-hop}  & 85.1 & 89.1 & 82.2 & 85.5\\ \hline
\end{tabular}
\end{center}
\caption{Comparison of different hops.}
\label{tab:my-table}
\end{table}

\subsection{Analysis on Different Hops}
In this work, we add the concepts within 2-hop when creating the curriculum to form the next stage concept set. In this section, we compare the 2-hop strategy with the hop strategies of other numbers on BERT\textsubscript{Medium}. We only report the performance from 1-hop to 4-hop given that observation of the results from 5-hop shows a negligible difference.

The results are shown in Table 8. It is observed that CCM\textsubscript{2-hop} outperforms the other strategies. We observe that CCM\textsubscript{1-hop} comprises too few concepts at the second stage, leading to the degradation of generalization performance. By contrast, CCM\textsubscript{3-hop} and CCM\textsubscript{4-hop} already comprise almost all concepts in the second stage, and thus fail to learn concepts progressively. 

\begin{table}
\begin{center}
\begin{tabular}{l@{}cccc}
\hline
     & MRPC & SST & STS & Avg.  \\ \hline
2 stage & 84.5 & 90.5 & 83.3 & 86.4 \\
3 stage & 85.1 & 90.7 & \textbf{84.4} & 86.7 \\
4 stage\textsubscript{(w/o warmup)}& 82.1 & 89.7 & 83.2 & 85.0 \\
4 stage &  \textbf{86.7} & \textbf{90.9} & 83.6 & \textbf{87.0} \\
5 stage & 85.7  & 90.5 & 83.8 & 86.6\\ \hline 

\end{tabular}
\end{center}
\caption{Comparison of numbers of curriculum stages.}
\label{tab:my-table}
\end{table}

\subsection{Number of Curriculum Stages}
Table 9 shows the results of the CCM experiments according to the different number of curriculum stages. In this experiment, we expand the concepts within 2-hops at every stage, and at the last stage of all curricula, the concept set includes all the concepts in the knowledge graph. Experimental results show the highest performance when the curriculum is composed of four stages. We find that few concepts are added after 4 stages, and speculate that this undermines curriculum effectiveness.

\section{Conclusion}
In this work, we have proposed CCM that masks easy concepts first and gradually masks related concepts to the previously masked ones for language model pre-training. With the help of our carefully designed linguistic difficulty criteria and curriculum construction, CCM can offer an effective masking schedule for MLM. The experimental results demonstrate that our curriculum improves convergence speeds and the generalization performance of MLM. 

\section{Limitations}
CCM has achieved impressive results in improving the efficiency of the LM pre-training, but some limitations need to be tackled in future work. First, we have not tested the efficacy of other knowledge graphs such as Wikidata. We believe that utilizing several knowledge graphs jointly would expand the coverage of concepts and relations. 

Furthermore, although we think all the relationship in the knowledge graph reflects the relevance between concepts, it is necessary to study the degree to which each relationship reflects the relevance between concepts. The concepts connected by specific relationships may benefit more from previous learning than concepts connected by others. We plan to construct a more sophisticated curriculum by handling each relationship differently.

\section{Acknowledgement}
We thank the anonymous reviewers for their helpful comments. This work was supported by the Basic Research Program through the National Research Foundation of Korea (NRF) grant funded by the Korea government (MSIT) (2020R1A4A1018309), National Research Foundation of Korea (NRF) grant funded by the Korea government (MSIT) (2021R1A2C3010430) and Institute of Information communications Technology Planning Evaluation (IITP) grant funded by the Korea government (MSIT) (No. 2019-0-00079, Artificial Intelligence Graduate School Program (Korea University)).

\bibliography{custom}

\begin{thebibliography}{37}
\expandafter\ifx\csname natexlab\endcsname\relax\def\natexlab#1{#1}\fi

\bibitem[{Anglin(1978)}]{Anglin-Child:1978}
Jeremy~M. Anglin. 1978.
\newblock From reference to meaning.
\newblock \emph{Child Development}, 49:969--976.

\bibitem[{Bengio et~al.(2009)Bengio, Louradour, Collobert, and
  Weston}]{Bengio-ICML:2009}
Yoshua Bengio, J{\'{e}}r{\^{o}}me Louradour, Ronan Collobert, and Jason Weston.
  2009.
\newblock Curriculum learning.
\newblock In \emph{Proceedings of the 26th Annual International Conference on
  Machine Learning}, pages 41--48.

\bibitem[{Campos(2021)}]{Campos-arXiv:2021}
Daniel Campos. 2021.
\newblock Curriculum learning for language modeling.
\newblock \emph{Computing Research Repository}, abs/2108.02170.

\bibitem[{Cer et~al.(2017)Cer, Diab, Agirre, Lopez{-}Gazpio, and
  Specia}]{Cer-SemEval:2017}
Daniel~M. Cer, Mona~T. Diab, Eneko Agirre, I{\~{n}}igo Lopez{-}Gazpio, and
  Lucia Specia. 2017.
\newblock Semeval-2017 task 1: Semantic textual similarity multilingual and
  crosslingual focused evaluation.
\newblock In \emph{Proceedings of the 11th International Workshop on Semantic
  Evaluation}, pages 1--14.

\bibitem[{Clark et~al.(2020)Clark, Luong, Le, and Manning}]{Clark-ICLR:2020}
Kevin Clark, Minh{-}Thang Luong, Quoc~V. Le, and Christopher~D. Manning. 2020.
\newblock {ELECTRA:} pre-training text encoders as discriminators rather than
  generators.
\newblock In \emph{8th International Conference on Learning Representations}.

\bibitem[{Collobert et~al.(2011)Collobert, Weston, Bottou, Karlen, Kavukcuoglu,
  and Kuksa}]{Collobert-jMach:2011}
Ronan Collobert, Jason Weston, L{\'{e}}on Bottou, Michael Karlen, Koray
  Kavukcuoglu, and Pavel~P. Kuksa. 2011.
\newblock Natural language processing (almost) from scratch.
\newblock \emph{J. Mach. Learn. Res.}, 12:2493--2537.

\bibitem[{Devlin et~al.(2019)Devlin, Chang, Lee, and
  Toutanova}]{Devlin-NAACL:2019}
Jacob Devlin, Ming{-}Wei Chang, Kenton Lee, and Kristina Toutanova. 2019.
\newblock {BERT:} pre-training of deep bidirectional transformers for language
  understanding.
\newblock In \emph{Proceedings of the 2019 Conference of the North American
  Chapter of the Association for Computational Linguistics: Human Language
  Technologies}, pages 4171--4186.

\bibitem[{Dolan and Brockett(2005)}]{Dolan-IWP@IJCNLP:2005}
William~B. Dolan and Chris Brockett. 2005.
\newblock Automatically constructing a corpus of sentential paraphrases.
\newblock In \emph{Proceedings of the Third International Workshop on
  Paraphrasing}.

\bibitem[{Elman(1993)}]{elman-Cognition:1993}
Jeffrey~L. Elman. 1993.
\newblock Learning and development in neural networks: the importance of
  starting small.
\newblock \emph{Cognition}, 48:71--99.

\bibitem[{Giampiccolo et~al.(2007)Giampiccolo, Magnini, Dagan, and
  Dolan}]{Giampiccolo-ACL-PASCAL:2007}
Danilo Giampiccolo, Bernardo Magnini, Ido Dagan, and Bill Dolan. 2007.
\newblock The third {PASCAL} recognizing textual entailment challenge.
\newblock In \emph{Proceedings of the ACL-PASCAL workshop on textual entailment
  and paraphrasing}, pages 1--9.

\bibitem[{Horton and Markman(1980)}]{Horton-Child:1980}
Marjorie~S. Horton and Ellen~M. Markman. 1980.
\newblock Developmental differences in the acquisition of basic and
  superordinate categories.
\newblock \emph{Child Development}, 51:708--719.

\bibitem[{Jiang et~al.(2020)Jiang, Yu, Zhou, Chen, Feng, and
  Yan}]{Jiang-NeurIPS:2020}
Zihang Jiang, Weihao Yu, Daquan Zhou, Yunpeng Chen, Jiashi Feng, and Shuicheng
  Yan. 2020.
\newblock Convbert: Improving {BERT} with span-based dynamic convolution.
\newblock In \emph{Advances in Neural Information Processing Systems 33}.

\bibitem[{Joshi et~al.(2020)Joshi, Chen, Liu, Weld, Zettlemoyer, and
  Levy}]{Joshi-TACL:2020}
Mandar Joshi, Danqi Chen, Yinhan Liu, Daniel~S. Weld, Luke Zettlemoyer, and
  Omer Levy. 2020.
\newblock Spanbert: Improving pre-training by representing and predicting
  spans.
\newblock \emph{Trans. Assoc. Comput. Linguistics}, 8:64--77.

\bibitem[{Krueger and Dayan(2009)}]{Krueger-Cognition:2009}
Kai~A. Krueger and Peter Dayan. 2009.
\newblock Flexible shaping: how learning in small steps helps.
\newblock \emph{Cognition}, 110:380--394.

\bibitem[{Kumar et~al.(2010)Kumar, Packer, and Koller}]{Kumar-NeurIPS:2010}
M.~Pawan Kumar, Benjamin Packer, and Daphne Koller. 2010.
\newblock Self-paced learning for latent variable models.
\newblock In \emph{Advances in Neural Information Processing Systems 23}, pages
  1189--1197.

\bibitem[{Levine et~al.(2021)Levine, Lenz, Lieber, Abend, Leyton{-}Brown,
  Tennenholtz, and Shoham}]{Levine-ICLR:2021}
Yoav Levine, Barak Lenz, Opher Lieber, Omri Abend, Kevin Leyton{-}Brown, Moshe
  Tennenholtz, and Yoav Shoham. 2021.
\newblock Pmi-masking: Principled masking of correlated spans.
\newblock In \emph{Proceedings of 9th International Conference on Learning
  Representations}.

\bibitem[{Lewis et~al.(2020)Lewis, Liu, Goyal, Ghazvininejad, Mohamed, Levy,
  Stoyanov, and Zettlemoyer}]{Lewis-ACL:2020}
Mike Lewis, Yinhan Liu, Naman Goyal, Marjan Ghazvininejad, Abdelrahman Mohamed,
  Omer Levy, Veselin Stoyanov, and Luke Zettlemoyer. 2020.
\newblock {BART:} denoising sequence-to-sequence pre-training for natural
  language generation, translation, and comprehension.
\newblock In \emph{Proceedings of the 58th Annual Meeting of the Association
  for Computational Linguistics}, pages 7871--7880.

\bibitem[{Nagatsuka et~al.(2021)Nagatsuka, Broni{-}Bediako, and
  Atsumi}]{Nagatsuka-RANLP:2021}
Koichi Nagatsuka, Clifford Broni{-}Bediako, and Masayasu Atsumi. 2021.
\newblock Pre-training a {BERT} with curriculum learning by increasing
  block-size of input text.
\newblock In \emph{Proceedings of the International Conference on Recent
  Advances in Natural Language Processing}, pages 989--996.

\bibitem[{Narayanan et~al.(2021)Narayanan, Shoeybi, Casper, LeGresley, Patwary,
  Korthikanti, Vainbrand, Kashinkunti, Bernauer, Catanzaro, Phanishayee, and
  Zaharia}]{Narayanan:2021}
Deepak Narayanan, Mohammad Shoeybi, Jared Casper, Patrick LeGresley, Mostofa
  Patwary, Vijay Korthikanti, Dmitri Vainbrand, Prethvi Kashinkunti, Julie
  Bernauer, Bryan Catanzaro, Amar Phanishayee, and Matei Zaharia. 2021.
\newblock Efficient large-scale language model training on {GPU} clusters using
  megatron-lm.
\newblock In \emph{Proceedings of the International Conference for High
  Performance Computing, Networking, Storage and Analysis}, pages 1--58.

\bibitem[{Pennington et~al.(2014)Pennington, Socher, and
  Manning}]{Pennington-EMNLP:2014}
Jeffrey Pennington, Richard Socher, and Christopher~D. Manning. 2014.
\newblock Glove: Global vectors for word representation.
\newblock In \emph{Proceedings of the 2014 Conference on Empirical Methods in
  Natural Language Processing}, pages 1532--1543.

\bibitem[{Qiu et~al.(2020)Qiu, Sun, Xu, Shao, Dai, and Huang}]{Qiu-CoRR:2020}
Xipeng Qiu, Tianxiang Sun, Yige Xu, Yunfan Shao, Ning Dai, and Xuanjing Huang.
  2020.
\newblock Pre-trained models for natural language processing: {A} survey.
\newblock \emph{CoRR}, abs/2003.08271.

\bibitem[{Rajpurkar et~al.(2016)Rajpurkar, Zhang, Lopyrev, and
  Liang}]{Rajpurkar-EMNLP:2016}
Pranav Rajpurkar, Jian Zhang, Konstantin Lopyrev, and Percy Liang. 2016.
\newblock Squad: 100,000+ questions for machine comprehension of text.
\newblock In \emph{Proceedings of the 2016 Conference on Empirical Methods in
  Natural Language Processing}, pages 2383--2392.

\bibitem[{Sachan and Xing(2016)}]{Sachan-ACL:2016}
Mrinmaya Sachan and Eric~P. Xing. 2016.
\newblock Easy questions first? {A} case study on curriculum learning for
  question answering.
\newblock In \emph{Proceedings of the 54th Annual Meeting of the Association
  for Computational Linguistics}, pages 7871--7880.

\bibitem[{Socher et~al.(2013)Socher, Perelygin, Wu, Chuang, Manning, Ng, and
  Potts}]{Socher-EMNLP:2013}
Richard Socher, Alex Perelygin, Jean Wu, Jason Chuang, Christopher~D. Manning,
  Andrew~Y. Ng, and Christopher Potts. 2013.
\newblock Recursive deep models for semantic compositionality over a sentiment
  treebank.
\newblock In \emph{Proceedings of the 2013 Conference on Empirical Methods in
  Natural Language Processing}, pages 1631--1642.

\bibitem[{Speer et~al.(2017)Speer, Chin, and Havasi}]{Speer-AAAI:2017}
Robyn Speer, Joshua Chin, and Catherine Havasi. 2017.
\newblock Conceptnet 5.5: An open multilingual graph of general knowledge.
\newblock In \emph{Proceedings of the Thirty-First {AAAI} Conference on
  Artificial Intelligence}, pages 4444--4451.

\bibitem[{Sun et~al.(2019)Sun, Wang, Li, Feng, Chen, Zhang, Tian, Zhu, Tian,
  and Wu}]{Sun-CoRR:2019}
Yu~Sun, Shuohuan Wang, Yu{-}Kun Li, Shikun Feng, Xuyi Chen, Han Zhang, Xin
  Tian, Danxiang Zhu, Hao Tian, and Hua Wu. 2019.
\newblock {ERNIE:} enhanced representation through knowledge integration.
\newblock \emph{CoRR}, abs/1904.09223.

\bibitem[{Tay et~al.(2019)Tay, Wang, Luu, Fu, Phan, Yuan, Rao, Hui, and
  Zhang}]{Tay-ACL:2019}
Yi~Tay, Shuohang Wang, Anh~Tuan Luu, Jie Fu, Minh~C. Phan, Xingdi Yuan, Jinfeng
  Rao, Siu~Cheung Hui, and Aston Zhang. 2019.
\newblock Simple and effective curriculum pointer-generator networks for
  reading comprehension over long narratives.
\newblock In \emph{Proceedings of the 57th Conference of the Association for
  Computationalx Linguistics}, pages 4922--4931.

\bibitem[{Wang et~al.(2019)Wang, Singh, Michael, Hill, Levy, and
  Bowman}]{Wang-ICLR:2019}
Alex Wang, Amanpreet Singh, Julian Michael, Felix Hill, Omer Levy, and
  Samuel~R. Bowman. 2019.
\newblock {GLUE:} {A} multi-task benchmark and analysis platform for natural
  language understanding.
\newblock In \emph{7th International Conference on Learning Representations}.

\bibitem[{Warstadt et~al.(2019)Warstadt, Singh, and
  Bowman}]{Warstadt-Trans:2019}
Alex Warstadt, Amanpreet Singh, and Samuel~R. Bowman. 2019.
\newblock Neural network acceptability judgments.
\newblock \emph{Trans. Assoc. Comput. Linguistics}, 7.

\bibitem[{Williams(1992)}]{Williams-Mach:1992}
Ronald~J. Williams. 1992.
\newblock Simple statistical gradient-following algorithms for connectionist
  reinforcement learning.
\newblock \emph{Mach. Learn.}, 8.

\bibitem[{Xu et~al.(2020)Xu, Zhang, Mao, Wang, Xie, and Zhang}]{Xu-ACL:2020}
Benfeng Xu, Licheng Zhang, Zhendong Mao, Quan Wang, Hongtao Xie, and Yongdong
  Zhang. 2020.
\newblock Curriculum learning for natural language understanding.
\newblock In \emph{Proceedings of the 58th Annual Meeting of the Association
  for Computational Linguistics}, pages 6095--6104.

\bibitem[{Yang et~al.(2019)Yang, Dai, Yang, Carbonell, Salakhutdinov, and
  Le}]{Yang-NeurIPS:2019}
Zhilin Yang, Zihang Dai, Yiming Yang, Jaime~G. Carbonell, Ruslan Salakhutdinov,
  and Quoc~V. Le. 2019.
\newblock Xlnet: Generalized autoregressive pretraining for language
  understanding.
\newblock In \emph{Advances in Neural Information Processing Systems 32}, pages
  5754--5764.

\bibitem[{Zhan et~al.(2021)Zhan, Liu, Wong, and Chao}]{Zhan-AAAI:2021}
Runzhe Zhan, Xuebo Liu, Derek~F. Wong, and Lidia~S. Chao. 2021.
\newblock Meta-curriculum learning for domain adaptation in neural machine
  translation.
\newblock In \emph{Proceedings of Thirty-Fifth {AAAI} Conference on Artificial
  Intelligence}, pages 14310--14318.

\bibitem[{Zhang et~al.(2018)Zhang, Kumar, Khayrallah, Murray, Gwinnup,
  Martindale, McNamee, Duh, and Carpuat}]{Zhang-CoRR:2018}
Xuan Zhang, Gaurav Kumar, Huda Khayrallah, Kenton Murray, Jeremy Gwinnup,
  Marianna~J. Martindale, Paul McNamee, Kevin Duh, and Marine Carpuat. 2018.
\newblock An empirical exploration of curriculum learning for neural machine
  translation.
\newblock \emph{CoRR}, abs/1811.00739.

\bibitem[{Zhang et~al.(2019)Zhang, Han, Liu, Jiang, Sun, and
  Liu}]{Zhang-ACL:2019}
Zhengyan Zhang, Xu~Han, Zhiyuan Liu, Xin Jiang, Maosong Sun, and Qun Liu. 2019.
\newblock {ERNIE:} enhanced language representation with informative entities.
\newblock In \emph{Proceedings of the 57th Conference of the Association for
  Computational Linguistics}, pages 1441--1451.

\bibitem[{Zhou et~al.(2020)Zhou, Yang, Wong, Wan, and Chao}]{Zhou-ACL:2020}
Yikai Zhou, Baosong Yang, Derek~F. Wong, Yu~Wan, and Lidia~S. Chao. 2020.
\newblock Uncertainty-aware curriculum learning for neural machine translation.
\newblock In \emph{Proceedings of the 58th Annual Meeting of the Association
  for Computational Linguistics}, pages 6934--6944.

\bibitem[{Zhu et~al.(2015)Zhu, Kiros, Zemel, Salakhutdinov, Urtasun, Torralba,
  and Fidler}]{Zhu-ICCV:2015}
Yukun Zhu, Ryan Kiros, Richard~S. Zemel, Ruslan Salakhutdinov, Raquel Urtasun,
  Antonio Torralba, and Sanja Fidler. 2015.
\newblock Aligning books and movies: Towards story-like visual explanations by
  watching movies and reading books.
\newblock In \emph{Proceedings of the {IEEE} International Conference on
  Computer Vision}, pages 19--27.

\end{thebibliography}
\bibliographystyle{acl_natbib}

\end{document}